\documentclass[10.7pt,]{article}

\usepackage[letterpaper, margin=2.54cm, top=2.54cm]{geometry}
\usepackage[super,comma,sort&compress]{natbib}
\usepackage{lmodern}
\usepackage{authblk} 
\usepackage{amssymb,amsmath}
\usepackage{wrapfig}
\usepackage{subfig} 
\usepackage{graphicx,grffile}
\usepackage{booktabs}
\usepackage[table,xcdraw]{xcolor}
\usepackage[labelfont=bf,labelsep=period]{caption}
\usepackage{ifxetex,ifluatex}
\ifnum 0\ifxetex 1\fi\ifluatex 1\fi=0 
  \usepackage[T1]{fontenc}
  \usepackage[utf8]{inputenc}
\else 
  \ifxetex
    \usepackage{mathspec}
  \else
    \usepackage{fontspec}
  \fi
  \defaultfontfeatures{Ligatures=TeX,Scale=MatchLowercase}
\fi
\IfFileExists{upquote.sty}{\usepackage{upquote}}{}
\IfFileExists{microtype.sty}{%
	\usepackage{microtype}
	\UseMicrotypeSet[protrusion]{basicmath} 
}{}

\usepackage{lipsum} 

\usepackage{sectsty}
\allsectionsfont{\fontsize{10}{10}\selectfont}
\subsectionfont{\itshape\bfseries\fontsize{10}{10}\selectfont}
\subsubsectionfont{\normalfont\itshape}

\makeatletter
\renewcommand\subsubsection{\@startsection{subsubsection}{3}{\z@}%
	{-3.25ex\@plus -1ex \@minus -.2ex}%
    {-1.5ex \@plus -.2ex}
    {\normalfont\itshape}}
\makeatother

\makeatletter 
\renewcommand\@biblabel[1]{#1.} 
\makeatother %

\usepackage{etoolbox}
\makeatletter
\patchcmd{\@maketitle}{\LARGE}{\bfseries\fontsize{15}{16}\selectfont}{}{}
\makeatother

\makeatletter
\def\maxwidth{\ifdim\Gin@nat@width>\linewidth\linewidth\else\Gin@nat@width\fi}
\def\maxheight{\ifdim\Gin@nat@height>\textheight\textheight\else\Gin@nat@height\fi}
\makeatother

\setkeys{Gin}{width=\maxwidth,height=\maxheight,keepaspectratio}
\ifx\paragraph\undefined\else
\let\oldparagraph\paragraph
\renewcommand{\paragraph}[1]{\oldparagraph{#1}\mbox{}}
\fi
\ifx\subparagraph\undefined\else
\let\oldsubparagraph\subparagraph
\renewcommand{\subparagraph}[1]{\oldsubparagraph{#1}\mbox{}}
\fi
\usepackage{hyperref}
\hypersetup{
	unicode=true,
	pdftitle={My Cool Title Here},
	pdfauthor={Author One, Author Two, Author Three},
	pdfkeywords={keyword1, keyword2},
	pdfborder={0 0 0},
	breaklinks=true
}
\title{\vspace{-2em} Opportunistic Screening for Pancreatic Cancer using Computed Tomography Imaging and Radiology Reports}
\author[ ]{\bf\fontsize{13}{14}\selectfont David Le, PhD\textsuperscript{1}, Ramon Correa-Medero, BS\textsuperscript{1}, Amara Tariq, PhD\textsuperscript{1}, Bhavik Patel, MD\textsuperscript{1}, Motoyo Yano, MD, PhD\textsuperscript{1}, Imon Banerjee, PhD\textsuperscript{1}\vspace{-.7em}}
\affil[1]{\bf\fontsize{13}{14}\selectfont Department of Radiology, Mayo Clinic, Phoenix, Arizona, USA}
\date{} 

\begin{document}
\maketitle
\vspace{-4em} 

\section{Abstract}\label{abstract}
\emph{Pancreatic ductal adenocarcinoma (PDAC) is a highly aggressive cancer, with most cases diagnosed at stage IV and a five-year overall survival rate below 5\%. Early detection and prognosis modeling are crucial for improving patient outcomes and guiding early intervention strategies. In this study, we developed and evaluated a deep learning fusion model that integrates radiology reports and CT imaging to predict PDAC risk. The model achieved a concordance index (C-index) of 0.6750 (95\% CI: 0.6429, 0.7121) and 0.6435 (95\% CI: 0.6055, 0.6789) on the internal and external dataset, respectively, for 5-year survival risk estimation. Kaplan-Meier analysis demonstrated significant separation (p<0.0001) between the low and high risk groups predicted by the fusion model. These findings highlight the potential of deep learning-based survival models in leveraging clinical and imaging data for pancreatic cancer.}

\section{Introduction}\label{introduction}
Pancreatic ductal adenocarcinoma (PDAC) is the most common type of pancreatic cancer and remains one of the deadliest malignancies worldwide. In 2020, there were 495,773 new PDAC cases, ranking it as the 14th most common cancer globally, while 466,003 deaths made it the 7th leading cause of cancer-related mortality \cite{rahib2021estimated}. By 2040, pancreatic cancer is projected to become the second leading cause of cancer-related death \cite{rahib2021estimated}. PDAC is categorized into four stages (I–IV), with significantly different survival rates: 83.7\%, 13.3\%, 4.2\%, and 1.3\% for stages I, II, III, and IV, respectively \cite{blackford2020recent}. However, most PDAC cases are diagnosed at advanced stages, with only 10.3\% of cases detected at stage I and 53.8\% diagnosed at stage IV, when the cancer has already metastasized \cite{blackford2024pancreatic}.

Late-stage diagnosis is largely due to the challenges associated with early detection of PDAC. In its early stages, PDAC is often asymptomatic, and when symptoms do occur, they are typically nonspecific (e.g., weight loss, nausea, abdominal bloating) and are often overlooked or mistaken for other, less serious conditions \cite{vareedayah2018pancreatic}. Imaging further complicates the diagnosis, as small tumors (<20 mm) can be isoattenuating on computed tomography (CT) scans, making them difficult to detect. Additionally, mass-forming pancreatitis and autoimmune pancreatitis can mimic PDAC on imaging, leading to potential misdiagnoses \cite{bilreiro2024imaging}. Given these challenges, risk estimation using survival analysis with multimodal data is crucial for improving early intervention. Predictive survival models can help stratify high-risk patients, guide clinical decision-making, and optimize screening efforts, ultimately improving PDAC outcomes.

The Cox proportional hazards (CPH) model is commonly used for survival analysis, particularly when leveraging quantitative parameter spaces \cite{fisher1999time}. However, the linear assumptions underlying the CPH model may limit its ability to capture complex survival patterns, thereby reducing its prognostic performance \cite{katzman2018deepsurv}. To address these limitations, recent studies have explored deep learning–based survival models. For example, Zhang et al. employed a convolutional neural network (CNN) trained on segmented pancreas CT images to predict survival probability \cite{zhang2020cnn}. Their model achieved a concordance index (C-index) of 0.651, demonstrating the potential of CNN-based approaches for survival prediction. Building on this work, Lee et al. proposed an ensemble model that integrates clinical and CT imaging data for predicting 2-year overall survival (OS) and 1-year recurrence-free survival (RFS), achieving an area under the curve (AUC) of 0.76 for 2-year OS and 0.74 for 1-year RFS, thus highlighting the benefit of combining multimodal data for survival analysis \cite{lee2022preoperative}. Similarly, Yao et al. introduced a multitask learning framework that leverages contrast-enhanced CT (CE-CT) imaging for the simultaneous prediction of OS and resection margin status for PDAC patients, achieving a C-index of 0.645 and demonstrating the feasibility of integrating imaging biomarkers for improved prognosis \cite{yao2021deepprognosis}.

While these studies have made significant strides in leveraging deep learning for PDAC survival prediction, they predominantly rely on data collected post-diagnosis. This approach, although valuable, limits opportunities for early intervention. Our study addresses this gap by integrating both textual and imaging data from patient records where the CT was ordered for other clinical assessments prior to a PDAC diagnosis (with a minimum lead time of one year). By analyzing pre-diagnostic data, our models aim to opportunistically identify high-risk individuals before clinical manifestation of the disease, thereby facilitating earlier detection and intervention strategies.

\section{Methods}\label{methods}
\begin{figure}[htb!]
    \centering
    \subfloat[Text only]{\includegraphics[width=0.3\textwidth]{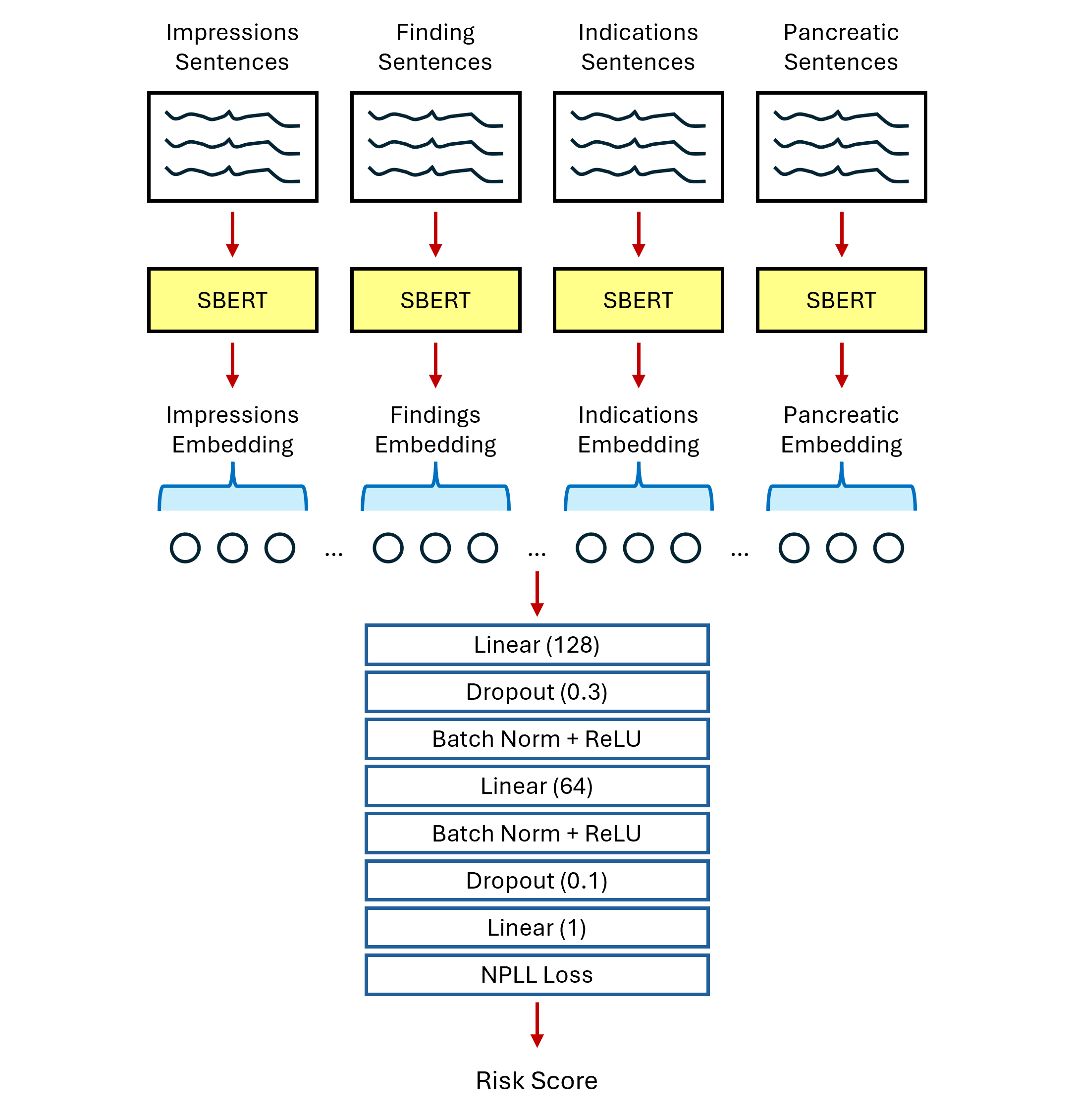}} 
    \subfloat[Image only]{\includegraphics[width=0.3\textwidth]{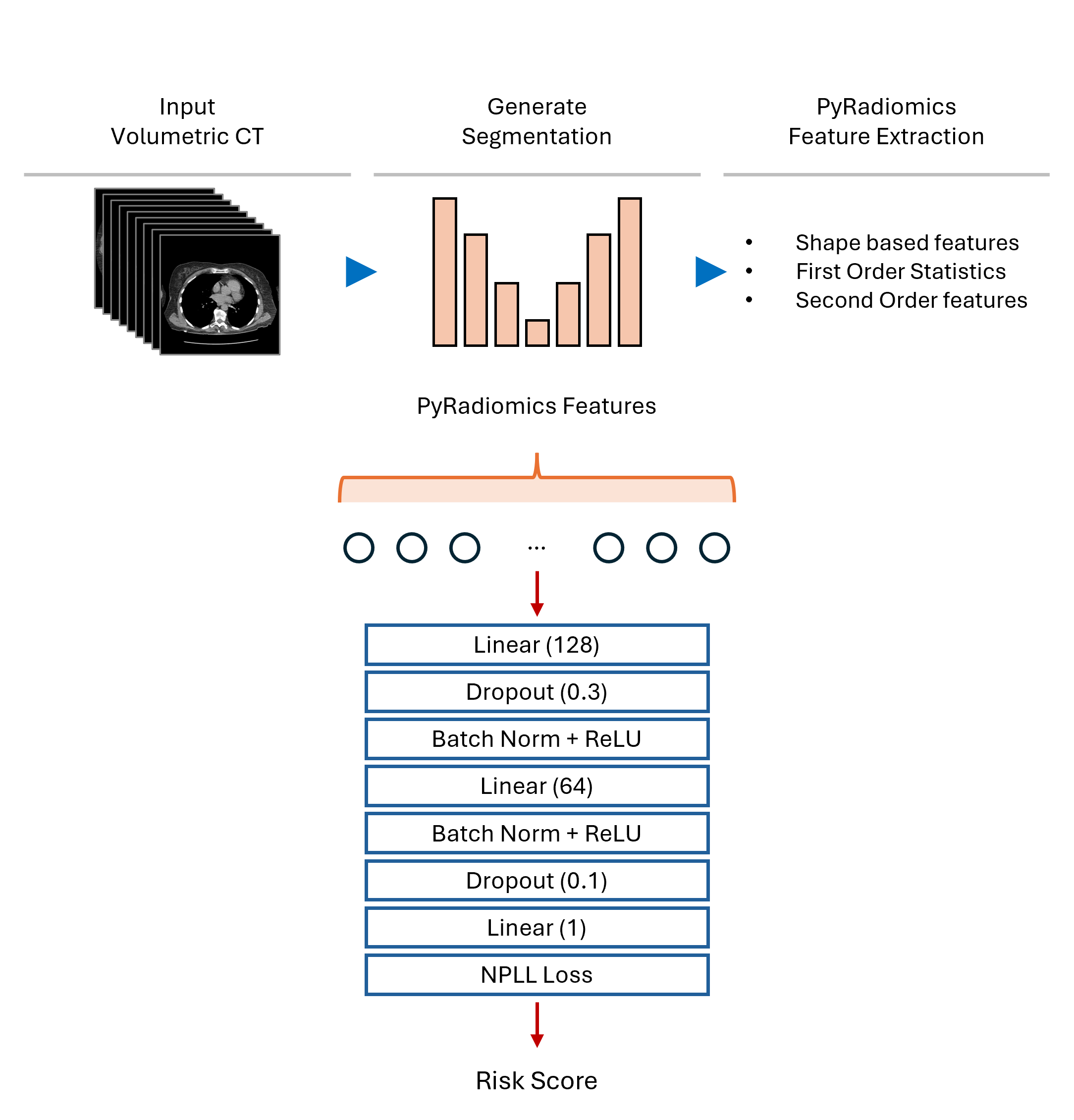}} 
    \subfloat[Fusion model]{\includegraphics[width=0.3\textwidth]{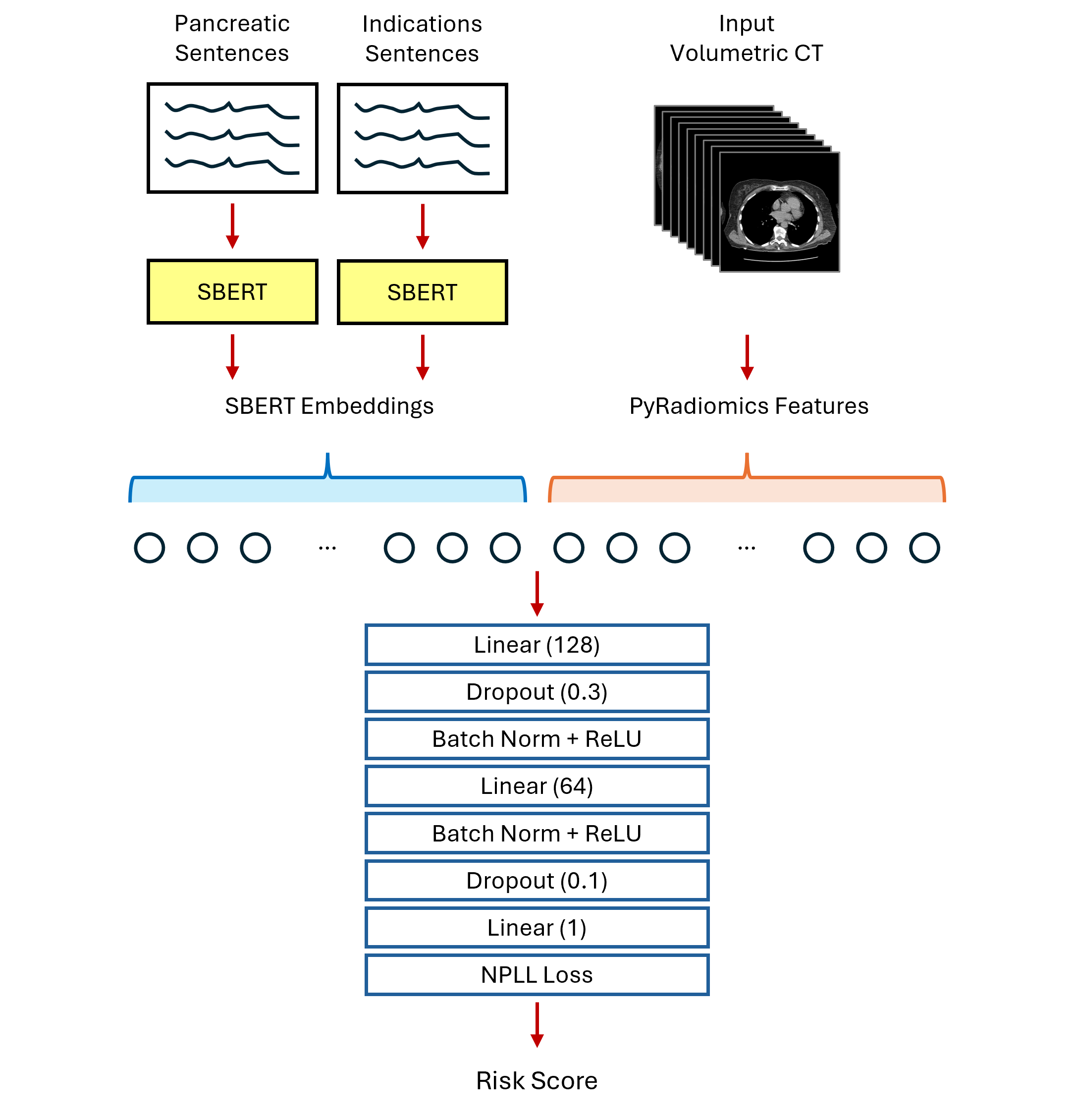}} 
    \caption{Pipeline for (a) text-only model, (b) image-only model, and (c) fusion model integrating text-based and CT volumetric information into a survival model. Sentence-BERT was used to extract embeddings from clinical report sentences, while pancreas segmentation was performed using TotalSegmentator, followed by feature extraction with PyRadiomics.}
    \label{fig:subfigures}
\end{figure}

The overall architecture of our proposed fusion model, along with separate models for individual modalities, is illustrated in Figure \ref{fig:subfigures}. The subsection below provides further details on each component.

\subsection{Text-Only Model: SBERT Embedding\label{sbert-embeddings}}

The text-only model was developed based on radiology reports corresponding to each pre-diagnosis CT scan. These reports, which cover the entire abdomen and often include specific radiological details about the pancreas, can contain valuable insights for predicting pancreatic cancer risk. However, the unstructured format and variability in style pose challenges for direct inclusion. Therefore, as a first step, we preprocessed the reports to extract relevant sentences. For example, the documents contained radiologist signatures mentioning names, dates, and various medical diagnostic codes, as well as non-standardized formatting (e.g., extra spaces and non-ASCII characters).

After cleaning the reports, we extracted sentences belonging to four categories: Findings, Impressions, Clinical Indications (i.e., reason for CT imaging), and Pancreatic characteristics. Additionally, we identified a separate category for sentences specifically related to the pancreas. If no sentences corresponded to these categories, we used a neutral placeholder sentence (e.g., “No recorded indications.” or “No significant findings noted.”). Once the unique sentences were derived for each category, we generated a 384-dimensional embedding per category using a pre-trained Sentence-BERT model \cite{reimers-2019-sentence-bert}. We then concatenated the embeddings from different sections (Figure~\ref{fig:subfigures}.a).

Next, we added a multilayer perceptron (MLP) on top of the concatenated embedding layer, incorporating batch normalization and dropout for regularization. The model was trained using the negative partial log-likelihood loss function until convergence on the validation split, with the goal of predicting a single 5-year risk score~\cite{kvamme2019time}.

\subsection{Image-Only Model: PyRadiomics + MLP}\label{pyradiomics-features}

In addition to textual information, we utilized volumetric CT imaging to model PDAC risk. For standardization, we filtered the abdominal CT scans for axial acquisitions and segmented the pancreas using TotalSegmentator, a tool for automatic segmentation of multiple anatomical structures from medical images \cite{wasserthal2023totalsegmentator, isensee2021nnu}. This pre-trained model can segment up to 104 different anatomical structures from medical imaging data. A recent study compared CT-based pancreatic segmentation models and found that TotalSegmentator demonstrated strong performance, achieving a mean Dice Similarity Coefficient (DSC) of 0.733 for normal pancreas cases and 0.703 for cases with pancreatitis, which is comparable to other models trained on significantly larger datasets \cite{somasundaram2024deep}.

Following pancreas segmentation, we extracted radiomic features using the PyRadiomics library, an open-source Python package that provides tools for the extraction of radiomic features from medical images~\cite{liu2019prediction}. We utilized the CT images and their corresponding segmentation to automatically derive features related to shape, intensity statistics, and texture from the 3D segmented pancreas \cite{van2017computational}. In total, 107 PyRadiomics features were extracted. These features were then fed into a multilayer perceptron (MLP) architecture similar to that used in our text-only model, with the model being trained using the negative partial log-likelihood loss function (Figure~\ref{fig:subfigures}.b).

\subsection{Multimodality Fusion Model}\label{multifeature-features}
After evaluating the two modalities separately, we combined the imaging and text-based features into a single model. Since the SBERT embeddings and PyRadiomics features are one-dimensional, we concatenated them and fed the result into a separate MLP model with end-to-end training (Figure~\ref{fig:subfigures}.c). We hypothesize that by applying early fusion of multimodal data—specifically combining medical images (such as CT) with radiology reports (textual findings)—the model can effectively learn the mutual information between these modalities. This fusion enables the model to correlate visual information from the images with the insights described in the reports, thereby enhancing its ability to predict risk scores more accurately. By integrating both modalities early in the model's training, we aim to capture richer, more comprehensive features, ultimately improving the predictive performance and robustness of the risk estimation system.

\subsection{Implementation}
In our study, all deep learning implementations were carried out using PyTorch (version 2.5.1) in Python (version 3.10.16). All deep survival models were optimized using the Adam optimizer with a learning rate of 1e-4 and weight decay of 1e-3, and were trained using the Negative Partial Log Likelihood (NPLL) loss function. The maximum number of epochs was set to 100, and early stopping was employed to prevent overfitting, with a patience of 10 epochs and a minimal delta of 1e-4. All models were trained on an NVIDIA RTX A5000 using CUDA version 12.4.

\section{Results}\label{results}
This retrospective study was approved by the Mayo Clinic Institutional Review Board. There was no patient or public involvement in the design, conduct, reporting, interpretation, or dissemination of the study. All cancer patients treated at Mayo Clinic between January 2011 and February 2023 at any of the three academic institution sites (Rochester, MN; Phoenix, AZ; and Jacksonville, FL) who underwent CT imaging prior to PDAC diagnosis were identified (n = 5,346). To ensure consistency, we applied selection criteria to include only patients with pre-diagnostic CT scans (for generic causes in the ED), where the CT scans were acquired in the axial orientation and were from the portal venous or later arterial phase. After filtering, a total of 2,530 patients met these criteria.

We designated patients from Jacksonville, FL, as the external test cohort, while patients from the other two institutions were further split into training and internal validation sets using an 80:20 train-validation ratio. The Jacksonville cohort differs from the training and validation cohorts not only in terms of geographical location and practice patterns, but also in gender and race (45.02\% female and 6.87\% multiracial), providing an evaluation of model performance under population shift. Since some patients had multiple CT scans prior to diagnosis, the dataset was split to ensure non-overlapping patients across the training, internal validation, and external validation sets. The final AI model development and validation cohort included 2,530 pancreatic cancer patients. The overall mean age was 67.62 ± 11.91 years, and 1,531 patients (58.97\%) were male. The average time between the CT scan and cancer diagnosis was 57.82 ± 51.33 months (median 44.50 months). For our analysis, patients diagnosed within the 5-year (60-month) interval were considered as events (uncensored), whereas those diagnosed after 60 months were treated as censored. Within this 5-year period, the censorship rate was 39.57\%. A detailed summary of patient demographics across the training, internal validation, and external validation cohorts is provided in Table~\ref{tab:patient-demographics}.

\begin{table}[htb!]
\centering
\caption{Patient demographic characteristics across training, validation and test split. Included 3 academic healthcare sites (Rochester MN, Phoenix AZ, and Jacksonville, FL)}
\label{tab:patient-demographics}
\resizebox{0.8\textwidth}{!}{
\begin{tabular}{@{}llll@{}}
\toprule
                                        & \textbf{Training} & \textbf{Validation} & \textbf{Test} \\ \midrule
Patients                       & 1744              & 364                 & 422              \\
Time Before Diagnosis (months, mean ± std) & 58.23 ± 51.41 & 56.78 ± 51.34 & 57.02 ± 51.13 \\
Time Before Diagnosis (years, mean ± std)  & 4.85 ± 4.28   & 4.73 ± 4.28   & 4.75 ± 4.26   \\
Censorship Rate                & 695 (39.85\%)     & 139 (38.19\%)       & 180 (42.65\%)    \\
Sites                                   & MN, AZ            & MN, AZ              & FL               \\
\textit{Age}                            &                   &                     &                  \\
\quad \textless 40                & 48 (2.75\%)   & 21 (5.76\%)   & 5 (1.18\%)    \\
\quad 40 - 60                        & 354 (20.29\%)     & 67 (18.4)           & 54 (12.79\%)     \\
\quad 60 - 80                        & 1087 (62.32\%)    & 221 (60.7\%)        & 320 (75.8\%)     \\
\quad \textgreater 80             & 255 (14.62\%) & 55 (15.1\%)   & 43 (10.18\%)  \\
\textit{Gender}                                                               \\
\quad Male                           & 1035 (59.34\%)    & 225 (61.81\%)       & 232 (54.9\%)     \\
\quad Female                         & 709 (40.6\%)      & 139 (38.1\%)        & 190 (45.02\%)    \\
\textit{Race}                                                                    \\
\quad White                          & 1671 (95.81\%)    & 347 (95.3\%)        & 351 (83.17\%)    \\
\quad Multi-race                     & 27 (1.54\%)       & 4 (1.09\%)          & 29 (6.87\%)      \\
\quad Other                          & 20 (1.14\%)       & 2 (0.55\%)          & 0 (0\%)          \\
\quad Black or African American                  & 15 (0.86\%)   & 6 (1.64\%)    & 42 (9.95\%)   \\
\quad American Indian/Alaskan Native & 6 (0.34\%)        & 0 (0\%)             & 0 (0\%)          \\
\quad Asian                          & 5 (0.28\%)        & 5 (1.37\%)          & 0 (0\%)          \\ \bottomrule
\end{tabular}
}
\end{table}

We used the C-index as the primary evaluation metric to assess model performance. The C-index measures the model’s ability to assign risk scores that align with survival times, where a C-index of 0.5 indicates random prediction and a C-index of 1.0 represents perfect prognostic accuracy. In this study, survival time was defined as the time to the first pancreatic cancer diagnosis. The best-performing model for PyRadiomics features utilizes all 107 features, whereas the best-performing SBERT embedding model employed Indications and Pancreas sentence embeddings. Therefore, we combined the best-performing features into our fusion model.

The results of our study indicate that our fusion model yields promising improvements in prognostic accuracy on the test set, as measured by the C-index. In particular, the combination of SBERT and PyRadiomics features achieved the highest performance, with an internal validation of 0.6750 (95\% CI: 0.6429, 0.7121) and an external validation of 0.6435 (95\% CI: 0.6055, 0.6789). These findings suggest that incorporating both clinical representations from SBERT and quantitative imaging features enhances the model's ability to predict the time to pancreatic cancer diagnosis more accurately than using PyRadiomics or clinical reports alone. This underscores the benefit of combining diverse data sources in prognostic approaches. The model evaluation metrics are summarized in Table~\ref{tab:main-result}.

\begin{table}[hbt!]
\centering
\caption{Best performing model evaluation metrics using Concordance Index for Internal and External datasets. 95\% confidence interval is calculated using bootstrapping. Optimal performance is highlighted in \textbf{bold}.}
\label{tab:main-result}
\resizebox{0.8\textwidth}{!}{%
\begin{tabular}{@{}lll@{}}
\toprule
Model                        & Internal Validation     & External Validation     \\ \midrule
Image only (PyRadiomics, no. features: 107) & 0.6430 (0.6012, 0.6801) & 0.5885 (0.5529, 0.6321) \\
Text only (Indications, Pancreas section)           & 0.6610 (0.6274, 0.6973) & 0.6410 (0.6088, 0.6750) \\
Fusion (SBERT and PyRadiomics )          & \textbf{0.6750 (0.6429, 0.7121)} & \textbf{0.6435 (0.6055, 0.6789)} \\ \bottomrule
\end{tabular}
}
\end{table}

\begin{figure}[htb!]
\centering
    \subfloat[Internal validation]{\includegraphics[width=\textwidth]{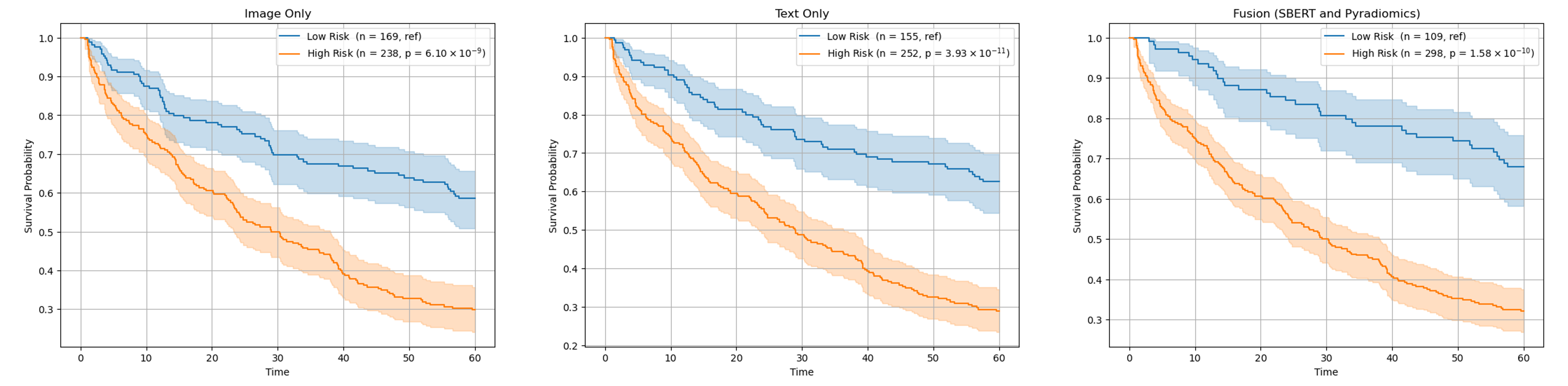}} \\
    \subfloat[External validation]{\includegraphics[width=\textwidth]{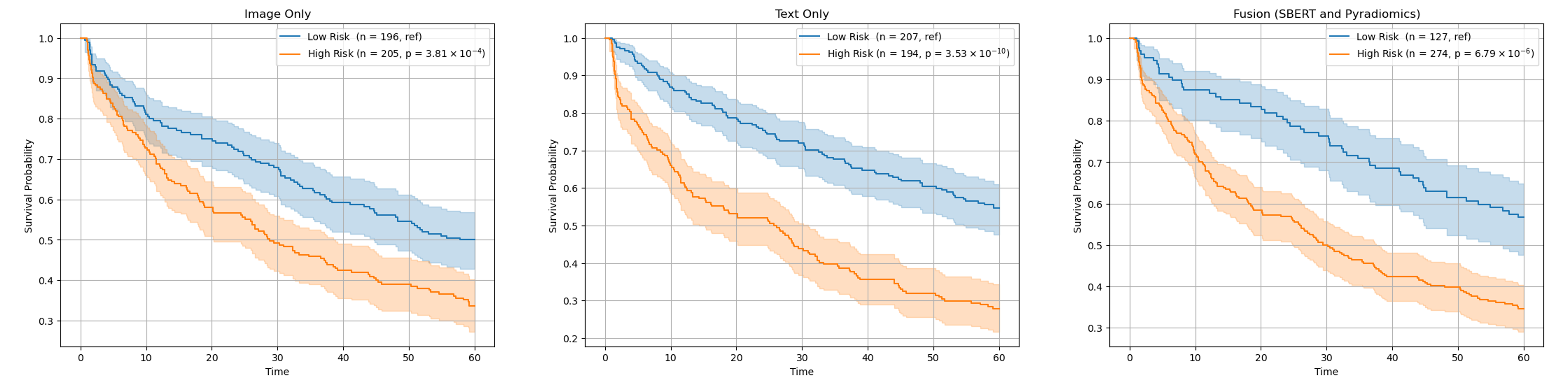}}
\caption{Kaplan-Meier curves for risk stratification between low and high risk for the image only, text only and fusion models for the internal and external validation dataset.}
\label{fig:km_curves}
\end{figure}

Furthermore, we evaluated each model’s capacity to stratify patients into high- and low-risk groups based on predicted risk scores, where values greater than 0 were designated as high risk and values less than or equal to 0 were designated as low risk. Kaplan-Meier curves for these stratified groups are presented in Figure \ref{fig:km_curves}, illustrating the estimated survival functions across both internal and external datasets. As shown, the image-only model using PyRadiomics features exhibited the weakest separation between high- and low-risk patients. In contrast, the text-only model provided improved differentiation, and the fusion model achieved the most pronounced distinction, particularly in the internal dataset, underscoring the benefit of integrating both textual and imaging information for risk stratification.

\subsection{Ablation Study}\label{sbert-comparison}
\begin{table}[hbt!]
\centering
\caption{Ablation study for survival models trained on different input features from PyRadiomics (image only) and SBERT sentence embeddings (text only). 95\% confidence interval is calculated using bootstrapping. Optimal performance is highlighted in \textbf{bold}.}
\label{tab:ablation-study}
\resizebox{0.8\textwidth}{!}{
\begin{tabular}{@{}lll@{}}
\toprule
Features                                   & Internal Validation     & External Validation     \\ \midrule
\textit{PyRadiomics}                                &                         &                         \\
\quad Num. features: 4, threshold \textless 0.1 & 0.5886 (0.5474,   0.6261) & 0.5359 (0.4994, 0.5750) \\
\quad Num. features: 7, threshold \textless 0.2  & 0.6130 (0.5710, 0.6486) & 0.5833 (0.5465, 0.6213) \\
\quad Num. features: 8, threshold \textless 0.3  & 0.6177 (0.5836, 0.6498) & 0.5681 (0.5341, 0.6036) \\
\quad Num. features: 12, threshold \textless 0.4 & 0.6277 (0.5939, 0.6654) & 0.5811 (0.5465, 0.6200) \\
\quad Num. features: 14, threshold \textless 0.5 & 0.6325 (0.6002, 0.6668) & 0.5838 (0.5428, 0.6178) \\
\quad Num. features: 21, threshold \textless 0.6 & 0.6364 (0.5986, 0.6732) & 0.5904 (0.5490, 0.6310) \\
\quad Num. features: 29, threshold \textless 0.7 & 0.6368 (0.5954, 0.6687) & 0.5943 (0.5625, 0.6332) \\
\quad Num. features: 41, threshold \textless 0.8 & 0.6299 (0.5998, 0.6607) & 0.5702 (0.5250, 0.6141) \\
\quad Num. features: 54, threshold \textless 0.9 & 0.6291 (0.5983, 0.6674) & 0.5734 (0.5346, 0.6101) \\
\quad Num. features: 107                         & \textbf{0.6430 (0.6012, 0.6801)} & \textbf{0.5885 (0.5529, 0.6321)} \\ \midrule
\textit{SBERT}                                      &                         &                         \\
\quad Impressions                                & 0.5271 (0.5118, 0.5472) & 0.5401 (0.5180, 0.5612) \\
\quad Findings                                   & 0.5809 (0.5475, 0.6135) & 0.5924 (0.5627, 0.6178) \\
\quad Pancreas                                   & 0.5803 (0.5459, 0.6135) & 0.6042 (0.5748, 0.6375) \\
\quad Indications                                & 0.6439 (0.6108, 0.6754) & 0.6392 (0.6073, 0.6762) \\
\quad Indications, Findings                      & 0.6528 (0.6171, 0.6890) & 0.6500 (0.6193, 0.6821) \\
\quad Indications, Impressions                   & 0.6600 (0.6245, 0.6906) & 0.6524 (0.6127, 0.6859) \\
\quad Indications, Pancreas                      & \textbf{0.6652 (0.6288, 0.6994)} & \textbf{0.6558 (0.6238, 0.6862)} \\
\quad Indications, Pancreas, Impressions         & 0.6652 (0.6268, 0.6994) & 0.6526 (0.6215, 0.6861) \\
\quad Indications, Pancreas, Findings            & 0.6538 (0.6165, 0.6888) & 0.6451 (0.6144, 0.6779) \\
\quad All sentence categories                    & 0.6578 (0.6243, 0.6887) & 0.6425 (0.6076, 0.6777) \\ \bottomrule
\end{tabular}
}
\end{table}

In our ablation study, we evaluated the prognostic impact of different feature subsets derived from PyRadiomics and SBERT sentence embeddings using internal and external validation. For the PyRadiomics models, we reduced feature redundancy by applying correlation thresholding. At a low threshold (0.1), only 4 features were retained, resulting in the poorest performance. As the threshold was relaxed to include more features (up to all 107), the model's performance improved steadily. The model incorporating all 107 PyRadiomics features achieved the highest performance with an internal C-index of 0.6430 (95\% CI: 0.6012, 0.6801) and an external C-index of 0.5885 (95\% CI: 0.5529, 0.6321), suggesting that a richer imaging feature set captures more relevant prognostic information.

For the text-embedding models, we investigated the predictive power of embeddings from various clinical report sections. Among the individual embeddings, the model based on the Indications section performed best (internal C-index = 0.6439 and external C-index = 0.6392). We then explored whether combining embeddings from multiple sections could further enhance performance. Our experiments revealed that the combination of Indications and Pancreas embeddings produced the highest prognostic accuracy, achieving an internal C-index of 0.6652 (95\% CI: 0.6288, 0.6994) and an external C-index of 0.6558 (95\% CI: 0.6238, 0.6862). Notably, adding further textual information (e.g., Findings or Impressions) did not yield additional benefits, indicating that the complementary information provided by clinical history and pancreatic findings (e.g., cysts, pancreatitis) is most critical for predicting the time to pancreatic cancer diagnosis.

Table~\ref{tab:ablation-study} summarizes the performance metrics across the different feature sets for both PyRadiomics and SBERT-based models. Overall, the ablation study underscores the importance of retaining a sufficiently large set of radiomic features and strategically combining clinical text embeddings to optimize prognostic performance.

\section{Conclusion}\label{conclusion}

We propose a novel opportunistic screening model for pancreatic cancer using a deep learning survival approach that integrates both textual and imaging data acquired one year prior to diagnosis. Through extensive experiments comparing different clinical report embeddings and PyRadiomics features derived from CT imaging, our study demonstrates the enhanced performance of a multimodal fusion model and highlights the importance of including complementary information in prognostic modeling. The model's effectiveness was validated on both internal and external datasets using the concordance index, which further emphasizes the potential for improving early risk stratification and guiding timely intervention in pancreatic cancer care through AI-driven risk score estimation. In future work, we will incorporate volumetric CT imaging and text reports into an end-to-end pipeline, eliminating the need for separate PyRadiomics feature extraction and allowing us to investigate the role of comprehensive imaging information, including torso fat, liver, and other relevant structures.

\section{Acknowledgements}\label{acknowledgements}
The project is supported by NCI funded `Multimodal Al Fusion Model for Early Detection for Pancreatic Cancer' (R01 CA289249-01).

\bibliographystyle{vancouver}
\bibliography{literature}

\end{document}